\def\year{2022}\relax
\documentclass[letterpaper]{article} 
\usepackage{aaai22}  
\usepackage{times}  
\usepackage{helvet}  
\usepackage{courier}  
\usepackage[hyphens]{url}  
\usepackage{graphicx} 
\usepackage{natbib}  
\usepackage{caption} 

\usepackage{bm}
\usepackage{amsfonts}
\DeclareCaptionStyle{ruled}{labelfont=normalfont,labelsep=colon,strut=off} 
\usepackage{algorithm}
\usepackage{algorithmic}

\usepackage{newfloat}
\usepackage{listings}
\floatstyle{ruled}
\newfloat{listing}{tb}{lst}{}
\floatname{listing}{Listing}
\title{Progressively Select and Reject Pseudo-labelled Samples\\ for Open-Set Domain Adaptation}
\author{
    Qian Wang\textsuperscript{\rm 1},
    Fanlin Meng\textsuperscript{\rm 2},
    Toby P. Breckon\textsuperscript{\rm 1,3}
}
\affiliations{
    \textsuperscript{\rm 1}Department of Computer Science, Durham University, UK\\
    \textsuperscript{\rm 2} Department of Mathmatical Sciences, University of Essex, UK\\
    \textsuperscript{\rm 3} Department of Engineering, Durham University, UK\\
    qian.wang173@hotmail.com
}

\begin{document}

\maketitle

\begin{abstract}
Domain adaptation solves image classification problems in the target domain by taking advantage of the labelled source data and unlabelled target data. Usually, the source and target domains share the same set of classes. As a special case, Open-Set Domain Adaptation (OSDA) assumes there exist additional classes in the target domain but not present in the source domain. To solve such a domain adaptation problem, our proposed method learns discriminative common subspaces for the source and target domains using a novel Open-Set Locality Preserving Projection (OSLPP) algorithm. The source and target domain data are aligned in the learned common spaces class-wisely. To handle the open-set classification problem, our method progressively selects target samples to be pseudo-labelled as known classes and rejects the outliers if they are detected as from unknown classes. The common subspace learning algorithm OSLPP simultaneously aligns the labelled source data and pseudo-labelled target data from known classes and pushes the rejected target data away from the known classes. The common subspace learning and the pseudo-labelled sample selection/rejection facilitate each other in an iterative learning framework and achieves state-of-the-art performance on benchmark datasets Office-31 and Office-Home with the average HOS of 87.4\% and 67.0\% respectively. 
\end{abstract}

\section{Introduction}
One key to the learning systems is the training data. The success of recent deep learning also relies on access to large-scale training data. Collecting and annotating a large amount of data for training can be difficult and costly in some domains \cite{wang2020generalized}. Given a target domain where the labelled training data are limited, one may promote the learning by exploiting annotated data from a source domain where annotated data are more easily to access but there exists a data distribution shift (i.e. domain shift) between the source and target domains. For example, object recognition from photos taken in the night as a target domain task can be better addressed by exploiting more easily accessible photos taken in the daytime (i.e. source domain). Domain adaptation aims at exploiting labelled source data to address the target-domain problem. Recently, a variety of approaches have been proposed to solve the closed set domain adaptation problems in which the source and target domain share the same set of classes \cite{chen2019progressive,chen2018joint}. In some real-world applications, however, there exist unlabelled target data from classes other than those shared ones. These data belong to \textbf{unknown} classes we are not interested in and are usually expected to be recognised as one unified class to discriminate them from known classes. This problem has been studied as the open set domain adaptation problem in literature \cite{saito2018open,liu2019separate,bucci2020effectiveness,luo2020progressive,fang2020open}.

Closed set domain adaptation methods suffer from the negative transfer issue \cite{wang2020unsupervised} when directly applied to the open set domain adaptation problems. Specifically, most closed set domain adaptation methods take advantage of the prior knowledge that the source and target domain share the same set of classes and conditional distributions can be well aligned. With the existence of unknown classes in the target domain for open-set domain adaptation, data belonging to these unknown classes will also be mistakenly aligned with some shared classes in the source domain. To solve this negative transfer issue of domain adaptation approaches, common strategies used in open-set recognition can be introduced and integrated into the domain adaptation frameworks towards improved performance for open set domain adaptation. One idea for this purpose is to detect the samples belonging to the unknown classes in the target domain and exclude them from the domain adaptation process so that the problem can be boiled down to a closed set domain adaptation problem. 

Following the same spirit, we propose a novel framework for open set domain adaptation based on the existing approach to closed set domain adaptation \cite{wang2020unsupervised}. The proposed framework has two novelties: the strategy of progressively selecting and rejecting pseudo-labelled samples and the domain adaptation method OSLPP (Open-Set Locality Preserving Projection), our modified version of the LPP algorithm for the open-set domain adaptation problem. In \cite{wang2020unsupervised}, the target domain data are pseudo-labelled as from the known classes and progressively selected as supervision for the domain adaptation process. In our proposed method, the target samples are also pseudo-labelled as the known classes and progressively selected as the supervision information for domain alignment. In addition, we also progressively reject some pseudo-labelled target samples as unknown classes if these samples are far away from all the known classes. As discussed in \cite{wang2020unsupervised}, it is important to select and reject the pseudo-labelled samples progressively so that the negative effect caused by incorrect pseudo-labelling can be mitigated.

Our proposed OSLPP aims at aligning the source and target domains in a learned subspace in which the source and target data from the known classes are aligned class-wisely whilst the target domain data pseudo-labelled as unknown classes are pushed away from the known classes. Inherited from the original LPP algorithm, the OSLPP has the capability of exploiting the data structure in the original feature space and is insensitive to the noise in the pseudo-labelling information. The OSLPP based subspace learning and pseudo-labelling are conducted alternately and repeated for a fixed number of iterations. As a result, these two components facilitate each other iteratively and the two domains are well aligned in the learned subspace where the recognition performance of target domain data is enhanced.

 The contributions of this work can be summarized as follows:
\begin{itemize}
    \item A novel framework is proposed for open set domain adaptation by learning a common subspace from both source and target domains using OSLPP, a novel algorithm aiming at aligning data from known classes and pushing away data from unknown classes.
    \item A strategy of progressively selecting and rejecting pseudo-labelled target domain data is proposed to facilitate the domain adaptation algorithm OSLPP.
    \item Experiments are conducted on two commonly used datasets Office-31 and Office-Home and the experimental results demonstrate our proposed method can achieve state-of-the-art performance.
\end{itemize}
\section{Related Work} \label{sec:related}
In this section, we review existing related work in domain adaptation (including closed set domain adaptation, partial domain adaptation and universal domain adaptation), open-set recognition and open-set domain adaptation.
\subsection{Domain Adaptation} \label{sec:da}
Domain adaptation is a general technique being used to address various research problems. From the perspective of tasks, it can be applied to image classification, object detection, image segmentation, etc. Based on the availability of labelled data in the target domain, there exist unsupervised, supervised and semi-supervised domain adaptation problems. One strong assumption in early domain adaptation research is that data from source and target domains share the same set of classes. In contrast to this closed-set domain adaptation, recently more practically useful formulations of domain adaptation problems have been studied including Partial domain adaption (PDA) \cite{wang2021source} where the source domain contains extra classes, OSDA which is our focus in this work and Universal Domain Adaptation \cite{you2019universal,saito2020universal,fu2020learning,li2021domain} where a versatile approach is expected to solve the problem without knowing it is a closed-set, open-set or partial domain adaptation problem,

Focusing on the image classification problem, the OSDA we focus on is unsupervised in the sense that there is no labelled target data. Our work is inspired by the unsupervised domain adaptation method proposed in \cite{wang2020unsupervised}. Using the same iterative learning framework, our method is dedicated to OSDA problems by detecting the unknown classes in the target domain and exploiting them to facilitate the common subspace learning for domain adaptation.

\subsection{Open-Set Domain Adaptation} \label{sec:openset-da}

Existing OSDA methods borrow the successful ideas from unsupervised domain adaptation approaches and adapt them for OSDA by handling the unknown target samples in specific ways. These methods are distinct from each other in how the unknown target samples are detected and utilized.  
\citet{saito2018open} learn a classifier to classify target samples into $C+1$ classes ($C$ known classes and 1 unknown class) and use such pseudo-labels to construct the loss function for classifier training. \citet{liu2019separate} take one step further by progressive adaptation with selected pseudo-labelled samples in the target domain. As this progressive learning strategy has been proved effective for domain adaptation \cite{wang2020unsupervised}, our work adopts a similar idea but implement it with completely different algorithms (i.e. data selection and rejection, domain alignment).  Following the same direction, \citet{fang2020open} use the samples classified as the unknown classes in the so-called \textit{open set difference} loss term to enhance the ability to recognize the unknown classes of the learned classifier. However, this method introduces too many hyper-parameters making it difficult to use in practice.
\citet{bucci2020effectiveness} address the OSDA problem in two stages. In the first stage, a classifier for separating known and unknown classes is learned using rotation-based self-supervised learning and in the second stage, the source samples together with the detected known target samples are combined to train a classifier classifying target samples into either one of the known classes or the unknown class. The limitation of this method is that the two stages are trained sequentially, hence the performance of the second stage relies on the accuracy of known-unknown class separability in the first stage and may lead to a sub-optimal solution.
\begin{figure*}[ht]
    \centering
    \includegraphics[width=0.95\textwidth]{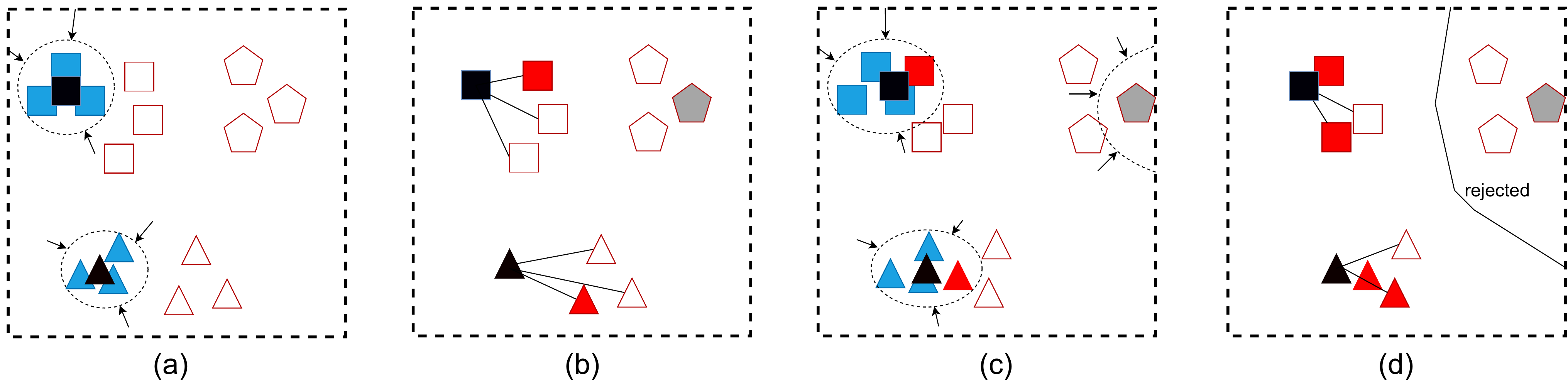}
    \caption{An illustration of our proposed approach to OSDA. The blue and red colours are used to represent the source and target domains, respectively; different shaped markers represent different classes; the black markers are the class means; the hollow red markers are unlabelled target samples which are filled in with the red colour if selected as known classes and filled in with the grey colour if rejected as unknown classes. (a) A supervised LPP is applied to the labelled source data to pull the source samples from the same class closer to each other. (b) In the subspace learned by (a), class means are computed and used to selectively assign pseudo-labels to the target samples; on the other hand, target samples far away from all class means are rejected as unknown classes. (c) Our proposed OSLPP is applied to the labelled source samples, selected pseudo-labelled target samples and rejected target samples so that the samples from the same class regardless of their domains are pulled close to each other whilst those rejected as unknown classes are pushed away from the known classes. (d) In the subspace learned by (c), class means are computed and used to selectively assign pseudo-labels to the known classes; the target samples are rejected if their nearest neighbour is already rejected. Steps (c) and (d) are repeated for $T$ times.}
    \label{fig:framework}
\end{figure*}

\citet{pan2020exploring} try to explore the structure of the target data by Self-Ensembling with Category-agnostic Clusters (SE-CC) to improve the recognition of unknown classes in the target domain.
\citet{kundu2020towards} aim at solving OSDA in a special scenario where source data are separated from the target data. Instead of training a model using the combination of source and target domain data, an inheritable model is firstly trained with the source data and subsequently adapted for the target data. The unknown classes are recognized by measuring the \textit{instance-level inheritability}. In addition, the samples confidently pseudo-labelled as from unknown classes are selected as target-domain supervision. Our work follows a similar idea of selecting the most confident pseudo-labels as supervision information from the target domain and such information is expected to better cluster the known classes and to push unknown classes far away from the known classes. The difference lies in that we use a novel manifold learning method OSLPP as opposed to the neural networks employed in \cite{kundu2020towards} and achieve superior performance.

\section{Problem Formulation} \label{sec:problem}
Suppose we have a labelled data set $\mathcal{D}^s = \{(\bm{x}^s_i,y^s_i)\}, i = 1,2,...,n_s$ from the source domain $\mathcal{S}$, $\bm{x}^s_i \in \mathbb{R}^{d_0}$ represents the feature vector of $i$-th labelled sample in the source domain, $d_0$ is the feature dimension and $y^s_i \in \mathcal{Y}^s$ denotes the corresponding label. OSDA aims at classifying an unlabelled data set $\mathcal{D}^t = \{\bm{x}^t_i\}, i=1,2,...,n_t$ from the target domain $\mathcal{T}$, where $\bm{x}^t_i \in \mathbb{R}^{d_0}$ represents the feature vector in the target domain. The target label space $\mathcal{Y}^t$ is a union of the source label space $\mathcal{Y}^s$ and an unknown label space $\mathcal{Y}^{unk}$ (i.e. $\mathcal{Y}^{unk} = \mathcal{Y}^{t} \backslash \mathcal{Y}^{s} \neq \emptyset $). The classes shared by the source and target domains are \textit{known classes} whilst the rest of target classes are unknown classes ($\mathcal{Y}^{unk}$). Samples from the target domain are expected to be classified as one of the known classes or the unified one unknown class. It is assumed that both the labelled source domain data $\mathcal{D}^s$ and the unlabelled target domain data $\mathcal{D}^t$ are available during model training.

\section{Method} \label{sec:method}
We introduce our proposed approach to the OSDA problems in this section. The framework of the approach is first described and illustrated in Figure \ref{fig:framework}. 
The original version of the LPP algorithm is briefly described to make the paper self-contained. 
Subsequently, we give the details of two key components in the framework: OSLPP and pseudo-label selection and rejection. Finally, the algorithm is summarized in Algorithm \ref{alg:oslpp}.

\subsection{Overview} \label{sec:overview}
Our approach to OSDA is inspired by the existing approaches to unsupervised domain adaptation \cite{wang2019unifying,chen2019progressive,wang2020unsupervised} and uses a similar iterative learning framework. The goal is to learn a common subspace based on the labelled source samples and pseudo-labelled target samples so that the source and target domains can be well aligned in the learned subspace and the target samples can be recognized by a simple nearest neighbour method. It has been proved the selective pseudo-labelling performs better than the methods considering all the pseudo-labelled target samples without the selection. The subspace learning procedure and the selective pseudo-labelling facilitate each other during the learning. Distinct from the existing works, we propose a new subspace learning method OSLPP, an extension of LPP to the OSDA problems, and a novel selection-rejection strategy to allow the OSLPP to learn a better subspace where known classes are aligned and unknown classes are pushed away.

\subsection{Locality Preserving Projection} \label{sec:lpp}
To make the paper self-contained, we briefly describe the original LPP algorithm \citet{he2004locality} based on which we propose our OSLPP in the next subsection. LPP aims at learning a favourable low-dimensional subspace where the local structures of data in the original feature space can be well preserved. Suppose $\bm{x}_i \in \mathbb{R}^{d_0}$ and $\bm{x}_j\in \mathbb{R}^{d_0}$ are two data points in the original feature space, LPP aims at learning a projection matrix $\bm{P} \in \mathbb{R}^{d_0\times d}$ ($d<<d_0$) so that data points close to each other in the original space will still be close in the projected subspace. The objective of LPP can be formulated as:
\begin{equation}
    \label{eq:lpp}
    \min_{\bm{P}} \sum_{i,j} ||\bm{P}^T \bm{x}_i - \bm{P}^T \bm{x}_j||_2^2 \bm{W}_{ij},
\end{equation}
where $\bm{W}$ is the similarity matrix of the graph constructed by all the data points. According to \cite{he2004locality}, the edges of the graph can be created by either $\epsilon-$neighbourhoods or $k$-nearest neighbours. The edge weights can be determined by the heat kernel $W_{ij} = e^{-\frac{||\bm{x}_i-\bm{x}_j||^2}{t}}$ or the simple binary assignment (i.e. all edges have the weights of 1).
Note that LPP is an unsupervised learning method without the need for labelling information. In the following subsection, we will describe how to extend the LPP algorithm to solve the OSDA problems where there exist unknown classes in the target domain.

\subsection{Open-Set LPP} \label{sec:oslpp}
Open set LPP aims at exploring the structural information underlying the labelled source data and the pseudo-labelled target data including those pseudo-labelled as known classes and unknown classes. The objective is the same as (\ref{eq:lpp}) but with a different way of constructing the similarity matrix $\bm{W} \in \mathbb{R}^{(n_s+n_t)\times(n_s+n_t)}$:
\begin{equation}
    \label{eq:w_oslpp}
    \mathbf{W}_{ij} = \left \{
    \begin{array}{ll}
    1, & y_i^s=y_j^s \lor y_i^s = \hat{y}_j^t\lor \hat{y}_i^t=y_i^s\lor \hat{y}_i^t = \hat{y}_j^t\\
    0 ,&  otherwise.
    \end{array}
    \right.
\end{equation}
where $y_i^s$ denotes the ground-truth label of $\bm{x}_i^s$ from the source domain, $\hat{y}_i^t$ denotes the pseudo-label of $\bm{x}_i^t$ from the target domain. It is noteworthy that $\hat{y}_i^t$ in (\ref{eq:w_oslpp}) can be either one of the known classes or the unified unknown class. Before the last iteration when all target samples are confidently recognised, only the confidently selected or rejected samples will be considered in (\ref{eq:w_oslpp}) and the rest of the target samples will be treated as uncertain samples and have 0 similarity with all the other samples.

By optimising the objective (\ref{eq:lpp}) with the similarity matrix define as (\ref{eq:w_oslpp}), the samples labelled or pseudo-labelled as the same class will be projected to be close to each other regardless of which domain they are from. The samples from different classes will be implicitly separated in the learned subspace. Target samples rejected as from unknown classes are treated as one unified class and hence will be pushed far away from all known classes implicitly. Adding terms to the objective explicitly pushing the rejected samples far away from known classes does not make a difference in our empirical study hence will not be discussed in this paper. 

According to \citep{he2004locality}, optimising the objective (\ref{eq:lpp}) is equivalent to solving the generalised eigenvalue problem:
\begin{equation}
\label{eq:eig}
\mathbf{X}\mathbf{D} \mathbf{X}^{T}\bm{p} = \lambda (\mathbf{X}\mathbf{L} \mathbf{X}^{T} + \mathbf{I}) \bm{p},
\end{equation}
where $\mathbf{L}=\mathbf{D}-\mathbf{W}$ is the laplacian matrix, $\mathbf{D}$ is a diagonal matrix with $\mathbf{D}_{ii}=\sum_j \mathbf{W}_{ij}$ and the regularization term $tr(\mathbf{P}^T \mathbf{P})$ is added for penalizing extreme values in the projection matrix $\mathbf{P}$.
Solving the generalized eigenvalue problem gives the optimal solution $\mathbf{P}=[\bm{p}_1, ..., \bm{p}_{d}]$ where $\bm{p}_1,...,\bm{p}_{d}$ are the eigenvectors corresponding to the largest $d$ eigenvalues.

In summary, the key of OSLPP is to treat the target samples differently when constructing the similarity matrix $\bm{W}$ based on if they are selected, rejected or uncertain. In the following subsection, we will describe how to make such decisions for the target samples.

\subsection{Pseudo-Labelled Sample Selection and Rejection}
Once the projection matrix $\bm{P}$ is learned, all the samples can be projected into the common subspace by:
\begin{equation}
    \label{eq:projection}
    \bm{z}^{s/t} = \bm{P}^T \bm{x}^{s/t}
\end{equation}
The pseudo labeling can be done in the subspace by the \textit{Nearest Class Mean} method \citep{wang2020unsupervised} for which the class means are computed over all the labelled source samples:
\begin{equation}
    \label{eq:classMean}
    \bm{\bar{z}}_c = \frac{1}{n_c} \sum_{y_i^s=c} \bm{z}_i^s
\end{equation}
and the pseudo label for a given target sample $\bm{x}^t$ can be predicted as:
\begin{equation}
    \label{eq:pseudolabel}
    \hat{y}^t = \arg\min_c dist(\bm{z}^t, \bm{\bar{z}}_c)
\end{equation}
with the probability:
\begin{equation}
    \label{eq:prob}
    p(\hat{y}^t=c) = \frac{e^{-dist(\bm{z}^t, \bm{\bar{z}}_c)}}{\sum_c e^{-dist(\bm{z}^t, \bm{\bar{z}}_c)}} 
\end{equation}
where $dist(\bm{a},\bm{b})$ is the Euclidean distance between $\bm{a}$ and $\bm{b}$.

In the $t$-th ($t=1,2,...,T$) iteration of learning, we select the top $(t+1)/T$ of the target samples pseudo-labelled as $c-$th class for each class $c=1,2,..., |\mathcal{C}|$. These selected samples together with their corresponding pseudo-labels will be used in the next iteration of subspace learning.

To reject target samples as from unknown classes, we use the 1-Nearest-Neighbour algorithm. A target sample is rejected if its nearest neighbour (excluding the samples neither selected nor rejected) in the subspace is a rejected sample. In the first iteration of learning, $n_r$ seed samples are rejected if their pseudo-labelling probabilities are top $n_r$ lowest, where $n_r$ is a hyper-parameter. The set of rejected samples will be enriched in each iteration. As a result, the value of $n_r$ will be a hyper-parameter trading off the accuracy of known classes and the unified unknown class.

We summarize the proposed approach in Algorithm \ref{alg:oslpp}.
\begin{algorithm}[tb]
	\caption{Open-Set LPP.}
	\label{alg:oslpp}
	\renewcommand{\algorithmicrequire}{\textbf{Input:}}
	\renewcommand{\algorithmicensure}{\textbf{Output:}}
	\begin{algorithmic}[1]
		\REQUIRE Labeled source data set $\mathcal{D}^s = \{(\bm{x}^s_i,y^s_i)\}, i = 1,2,...,n_s$ and unlabeled target data set $\mathcal{D}^{t}=\{\bm{x}_i^t\},i=1,2,...,n_{t}$, dimensionality of the subspace $d$, number of iteration $T$, number of initial rejected target samples $n_r$.
		\ENSURE The projection matrix $\mathbf{P}$ and predicted labels $\{\hat{y}^t\}$ for target samples.
		\STATE Initialize $k=0$;
		\STATE Learn the projection $\mathbf{P}_0$ using only source data $\mathcal{D}^s$;
		\STATE Assign pseudo labels for all target data using Eq. (\ref{eq:pseudolabel}-\ref{eq:prob});
		\STATE Initialize $n_r$ rejected samples of the lowest pseudo-labeling probabilities;
		\WHILE {$k < T$}
		\STATE $k \leftarrow k+1$;
		\STATE Select a subset of pseudo-labeled target data $\mathcal{S}_k \subset \mathcal{D}^t $;
		\STATE Reject samples using 1-Nearest-Neighbour and enrich the subset of rejected samples $\mathcal{R}_k \subset \mathcal{D}^t$;
		\STATE Learn $\bm{P}_k$ using $\mathcal{D}^s$, $\mathcal{S}_k$ and $\mathcal{R}_k$;
		\STATE Update pseudo labels for all target data using Eq.(\ref{eq:pseudolabel}).
		\ENDWHILE
	\end{algorithmic}
\end{algorithm}
\section{Experiments and Results} \label{sec:exp}
In this section, we demonstrate the experiments and results for validating the effectiveness of the proposed method. Specifically, we introduce the datasets, experimental settings, evaluation metrics and experimental results.

\subsection{Datasets} \label{sec:dataset}
Two commonly used datasets for OSDA were employed in our experiments: Office-31 and Office-Home. 
\textbf{Office31} \cite{saenko2010adapting} consists of three domains: Amazon (A), Webcam (W) and DSLR (D). There are 31 common classes for all three domains containing 4,110 images in total. Following the open set protocol employed in \cite{saito2018open, bucci2020effectiveness}, we use the first 10 classes in alphabetic order as the shared known classes in both source and target domains and the last 11 classes as the unknown classes in the target domain. Image features are extracted by the ResNet50 \cite{he2016deep} model pre-trained on ImageNet \cite{deng2009imagenet} without fine-tuning on the Office31 dataset.
\textbf{Office-Home} \cite{venkateswara2017deep} consists of four different domains: Artistic images (Ar), Clipart (Cl), Product images (Pr) and Real-World images (Rw). There are 65 object classes in each domain with a total number of 15,588 images. We follow \cite{bucci2020effectiveness} and use images from the first 25 classes in alphabetical order as the shared unknown classes in both domains and images from the remaining 40 classes as unknown classes in the target domain. Image features are extracted by the ResNet50 \cite{he2016deep} model pre-trained on ImageNet \cite{deng2009imagenet} without fine-tuning on the Office-Home dataset.

\subsection{Implementation details} \label{sec:implementation}
We implement the proposed method in MATLAB R2020b \footnote{Code will be released to the public.}. The 2048-dimensional ResNet50 features are first $l$2 normalised \cite{wang2020unsupervised} and the dimensionality is reduced by PCA\cite{wold1987principal} to 16 and 512 for Office31 and Office-Home datasets, respectively. Applying PCA to the original high-dimensional deep features not only can reduce the computation cost of OSLPP, but also benefits the performance in our experiments. The dimensionality $d$ of the learned subspace is set as 16 and 128 respectively. The number of iterations is set to 10 for both datasets. The number of initially rejected samples $n_r$ is set to 140 and 1200 respectively. The sensitivity of our method to these hyper-parameters will be discussed later in this section.

\subsection{Evaluation metrics} \label{sec:metrics}
There exist several evaluation metrics for evaluating OSDA approaches. \textbf{OS} is the mean per-class accuracy over all target domain images from known classes and unknown classes as one unified class. To measure the capabilities of recognizing the known and unknown classes, we use the metrics \textbf{OS*} which is the mean per-class accuracy over the shared known classes and \textbf{UNK} which is the accuracy of images from the unknown classes (as one unified class).
\begin{equation}
    \label{eq:os}
    OS = \frac{1}{|\mathcal{C}_s|+1} \sum_{i=1}^{|\mathcal{C}_s|+1} \frac{ |x: x \in \mathcal{D}_t^i \wedge \hat{y}(x)=i|}{|x: x \in \mathcal{D}_t^i|}
\end{equation}
\begin{equation}
    \label{eq:osstar}
    OS^* = \frac{1}{|\mathcal{C}_s|} \sum_{i=1}^{|\mathcal{C}_s} \frac{ |x: x \in \mathcal{D}_t^i \wedge \hat{y}(x)=i|}{|x: x \in \mathcal{D}_t^i|}
\end{equation}
\begin{equation}
    \label{eq:unk}
    UNK = \frac{|x: x \in \mathcal{D}_t^{unk} \wedge \hat{y}(x)=unk|}{|x: x \in \mathcal{D}_t^{unk}|}
\end{equation}
where $|\mathcal{C}_s|$ is the number of source domain classes (i.e. the number of shared known classes) in the class space $\mathcal{C}_s$, $\mathcal{D}_t^i$ denotes the data set of $i$-th class in the target domain and $\hat{y}(x)$ is the predicted label of data sample $x$.

OS is a combination of OS* and UNK as $OS = \frac{|\mathcal{C}_s|}{|\mathcal{C}_s|+1}\times OS^* + \frac{1}{|\mathcal{C}_s|+1}\times UNK$, however, as pointed out in \cite{bucci2020effectiveness}, OS can be dominated by the accuracy of known classes since the unknown classes are treated as one unified class. One effective evaluation metric properly balancing the recognition performance for known and unknown classes is the harmonic mean of OS* and UNK:
\begin{equation}
    \label{eq:hos}
    HOS = \frac{2\times OS^*\times UNK}{OS^* + UNK}
\end{equation}
In our experiments, we report OS*, UNK and HOS for individual domain adaptation tasks as well as the average performance over all possible tasks for a dataset.

\begin{table*}[!t]
	\centering
	{
		\centering
		\caption[]{Open-set Classification Accuracy (\%) on Office31 dataset using either ResNet50 features or ResNet50 based deep models ($\dagger$ indicates AlexNet was used).
		}
		\label{table:office31}
		\resizebox{2\columnwidth}{!}{%
			\begin{tabular}{l|ccc|ccc|ccc|ccc|ccc|ccc|ccc}
				\hline
				 & \multicolumn{3}{c|}{A$\to$D} & \multicolumn{3}{c|}{A$\to$W} & \multicolumn{3}{c|}{D$\to$A} & \multicolumn{3}{c|}{D$\to$W} & \multicolumn{3}{c|}{W$\to$A} & \multicolumn{3}{c|}{W$\to$D} & \multicolumn{3}{c}{Average} \\ 
				Method & OS* & UNK & HOS& OS* & UNK & HOS& OS* & UNK & HOS& OS* & UNK & HOS& OS* & UNK & HOS& OS* & UNK & HOS& OS* & UNK & HOS\\ \hline
				STA$_{sum}$ \cite{liu2019separate} &95.4 & 45.5 & 61.6 & 92.1 & 58.0 & 71.0 & 94.1 & 55.0 & 69.4 & 97.1 & 49.7 & 65.5 & 92.1 & 46.2 & 60.9 & 96.6 & 48.5 & 64.4 & 94.6 & 50.5 & 65.5\\
				STA$_{max}$ \cite{liu2019separate} & 91.0 & 63.9 & 75.0 & 86.7 & 67.6 & 75.9 & 83.1 & 65.9 & 73.2 & 94.1 & 55.5 & 69.8 & 66.2 & 68.0 & 66.1 & 84.9 & 67.8 & 75.2 & 84.3 & 64.8 & 72.6\\
				OSBP \cite{saito2018open} & 90.5 & 75.5 & 82.4 & 86.8 & 79.2 & 82.7 & 76.1 & 72.3 & 75.1 & 97.7 & 96.7 & $\bf97.2$ & 73.0 & 74.4 & 73.7 & 99.1 & 84.2 & 91.1 & 87.2 & 80.4 & 83.7\\
				UAN \cite{you2019universal} & 95.6 & 24.4 & 38.9 & 95.5 & 31.0 & 46.8 & 93.5 & 53.4 & 68.0 & 99.8 & 52.5 & 68.8 & 94.1 & 38.8 & 54.9 & 81.5 & 41.4 & 53.0 & 93.4 & 40.3 & 55.1 \\
				SE-CC$\dagger$ \cite{pan2020exploring} & 84.0 & 46.6 & 59.9 & 84.2 & 64.4 & 73.0 & 90.3 & 12.2 & 21.5 & 96.6 & 55.9 & 70.8 & 85.9 & 50.7 & 63.8 & 99.1 & 73.8 & 84.6 & 90.0 & 50.6 & 62.3 \\
				ROS \cite{bucci2020effectiveness} & 87.5 & 77.8 & 82.4 & 88.4 & 76.7 & 82.1 & 74.8 & 81.2 & 77.9 & 99.3 & 93.0 & 96.0 & 69.7 & 86.6 & 77.2 & 100.0 & 99.4 & $\bf99.7$ & 86.6 & 85.8 & 85.9\\
				OSLPP (Ours) & 92.6 & 90.4 & $\bf91.5$ & 89.5 & 88.4 & $\bf89.0$ & 82.1 & 76.6 & $\bf79.3$ & 96.9 & 88.0 & 92.3& 78.9 & 78.5 & \bf78.7& 95.8 & 91.5 & 93.6& 89.3 & 85.6 & $\bf87.4$ \\
				\hline
			\end{tabular}%
		}
	}
\end{table*}

\begin{table*}[!t]
	\centering
	{
		\centering
		\caption[]{Open-set Classification Accuracy (\%) on Office-Home dataset using either ResNet50 features or ResNet50 based deep models.
		}
		\label{table:officehome}
		\resizebox{2\columnwidth}{!}{%
			\begin{tabular}{l|ccc|ccc|ccc|ccc|ccc|ccc|ccc}
				\hline
				 & \multicolumn{3}{c|}{Ar$\to$Cl} & \multicolumn{3}{c|}{Ar$\to$Pr} & \multicolumn{3}{c|}{Ar$\to$Rw} & \multicolumn{3}{c|}{Cl$\to$Ar} & \multicolumn{3}{c|}{Cl$\to$Pr} & \multicolumn{3}{c|}{Cl$\to$Rw} \\ 
				Method & OS* & UNK & HOS& OS* & UNK & HOS& OS* & UNK & HOS& OS* & UNK & HOS& OS* & UNK & HOS& OS* & UNK & HOS\\ \cline{1-19}
				STA$_{sum}$ \cite{liu2019separate} & 50.8 & 63.4 & 56.3 & 68.7 & 59.7 & 63.7 & 81.1 & 50.5 & 62.1 & 53.0 & 63.9 & 57.9 & 61.4 & 63.5 & 62.5 & 69.8 & 63.2 & 66.3 \\
				STA$_{max}$ \cite{liu2019separate} & 46.0 & 72.3 & 55.8 & 68.0 & 48.4 & 54.0 & 78.6 & 60.4 & 68.3 & 51.4 & 65.0 & 57.4 & 61.8 & 59.1 & 60.4 & 67.0 & 66.7 & 66.8\\
				OSBP \cite{saito2018open} & 50.2 & 61.1 & 55.1 & 71.8 & 59.8 & 65.2 & 79.3 & 67.5 & 72.9 & 59.4 & 70.3 & \bf64.3 & 67.0 & 62.7 & 64.7 & 72.0 & 69.2 & \bf70.6\\
				UAN \cite{you2019universal} & 62.4 & 0.0 & 0.0 & 81.1 & 0.0 & 0.0 & 88.2 & 0.1 & 0.2 & 70.5 & 0.0 & 0.0 & 74.0 & 0.1 & 0.2 & 80.6 & 0.1 & 0.2 \\
				ROS \cite{bucci2020effectiveness} & 50.6 & 74.1 & 60.1 & 68.4 & 70.3 & 69.3 & 75.8 & 77.2 & \bf76.5 & 53.6 & 65.5 & 58.9 & 59.8 & 71.6 & 65.2 & 65.3 & 72.2 & 68.6 \\
				DAOD \cite{fang2020open} & 72.6 & 51.8 & 60.5 & 55.3 & 57.9 & 56.6 & 78.2 & 62.6 & 69.5 & 59.1 & 61.7 & 60.4 & 70.8 & 52.6 & 60.4 & 77.8 & 57.0 & 65.8 \\
				PGL \cite{luo2020progressive} & 63.3 & 19.1 & 29.3 & 78.9 & 32.1 & 45.6 & 87.7 & 40.9 & 55.8 & 85.9 & 5.3 & 10.0 & 73.9 & 24.5 & 36.8 & 70.2 & 33.8 & 45.6\\
				OSLPP (Ours) & 55.9 & 67.1 & \bf61.0& 72.5 & 73.1 & \bf72.8& 80.1 & 69.4 & 74.3& 49.6 & 79.0 & 60.9& 61.6 & 73.3 & \bf66.9& 67.2 & 73.9 & 70.4 \\

				\hline
				\hline

				 & \multicolumn{3}{c|}{Pr$\to$Ar} & \multicolumn{3}{c|}{Pr$\to$Cl} & \multicolumn{3}{c|}{Pr$\to$Rw} & \multicolumn{3}{c|}{Rw$\to$Ar} & \multicolumn{3}{c|}{Rw$\to$Cl} & \multicolumn{3}{c|}{Rw$\to$Pr} & \multicolumn{3}{c}{Average} \\ 
				Method & OS* & UNK & HOS& OS* & UNK & HOS& OS* & UNK & HOS& OS* & UNK & HOS& OS* & UNK & HOS& OS* & UNK & HOS& OS* & UNK & HOS\\ \hline
				STA$_{sum}$ \cite{liu2019separate} & 55.4 & 73.7 & 63.1 & 44.7 & 71.5 & 55.0 & 78.1 & 63.3 & 69.7 & 67.9 & 62.3 & 65.0 & 51.4 & 57.9 & 54.2 & 77.9 & 58.0 & 66.4 & 63.4 & 62.6 & 61.9\\
				STA$_{max}$ \cite{liu2019separate} & 54.2 & 72.4 & 61.9 & 44.2 & 67.1 & 53.2 & 76.2 & 64.3 & 69.5 & 67.5 & 66.7 & 67.1 & 49.9 & 61.1 & 54.5 & 77.1 & 55.4 & 64.5 & 61.8 & 63.3 & 61.1 \\
				OSBP \cite{saito2018open} & 59.1 & 68.1 & 63.2 & 44.5 & 66.3 & 53.2 & 76.2 & 71.7 & 73.9 & 66.1 & 67.3 & 66.7 & 48.0 & 63.0 & 54.5 & 76.3 & 68.6 & 72.3 & 64.1 & 66.3 & 64.7\\
				UAN \cite{you2019universal} & 73.7 & 0.0 & 0.0 & 59.1 & 0.0 & 0.0 & 84.0 & 0.1 & 0.2 & 77.5 & 0.1 & 0.2 & 66.2 & 0.0 & 0.0 & 85.0 & 0.1 & 0.1 &  75.2 & 0.0 & 0.1 \\
				ROS \cite{bucci2020effectiveness} & 57.3 & 64.3 & 60.6 & 46.5 & 71.2 & 56.3 & 70.8 & 78.4 & \bf74.4 & 67.0 & 70.8 & 68.8 & 51.5 & 73.0 & \bf60.4 & 72.0 & 80.0 & \bf75.7 & 61.6 & 72.4 & 66.2 \\
				DAOD \cite{fang2020open} & 71.3 & 50.5 & 59.1 & 58.4 & 42.8 & 49.4 & 81.8 & 50.6 & 62.5 & 66.7 & 43.3 & 52.5 & 60.0 & 36.6 & 45.5 & 84.1 & 34.7 & 49.1 & 69.6 & 50.2 & 57.6 \\
				PGL \cite{luo2020progressive} & 73.7 & 34.7 & 47.2 & 59.2 & 38.4 & 46.6 & 84.8 & 27.6 & 41.6 & 81.5 & 6.1 & 11.4 & 68.8 & 0.0 & 0.0 & 84.8 & 38.0 & 52.5 & 76.1 & 25.0 & 35.2\\
				OSLPP (Ours) & 54.6 & 76.2 & \bf63.6& 53.1 & 67.1 & \bf59.3& 77.0 & 71.2 & 74.0& 60.8 & 75.0 & \bf67.2& 54.4 & 64.3 & 59.0& 78.4 & 70.8 & 74.4& 63.8 & 71.7 & \bf67.0 \\

				\hline
			\end{tabular}%
		}
	}
\end{table*}

\subsection{Comparison with state of the art} \label{sec:results}
The proposed method OSLPP is compared against recent state-of-the-art approaches to OSDA problems. The approaches we compare against include STA  \cite{liu2019separate}, OSBP \cite{saito2018open}, UAN \cite{you2019universal}, ROS \cite{bucci2020effectiveness}, DAOD \cite{fang2020open} and PGL \cite{luo2020progressive}. We do not compare with \cite{panareda2017open} and \cite{pan2020exploring} since they either use a different experimental protocol or different deep features in their experiments and make it difficult for a direct comparison with those considered in our work.


To make a fair and reliable comparison, we report the results of state-of-the-art methods reproduced by \citet{bucci2020effectiveness} for STA, OSBP, UAN and ROS. For DAOD and PGL, the metrics UNK and HOS are calculated by Eqs. (\ref{eq:os}-\ref{eq:hos}) based on the reported OS and OS* values in their original papers. 

As we can see in Table \ref{table:office31}, our proposed method achieves the best average HOS of 87.4\% followed by 85.9\% by \citet{bucci2020effectiveness} and 83.7\% by \citet{saito2018open}. Among the six adaptation tasks, our method performs the best in terms of HOS on four of them. The other methods usually perform well in recognising the known classes (high OS*) but are bad at recognising the unknown classes (low UNK). This is partially due to the bias caused by only considering the evaluation metrics of OS and OS*. A high OS* along with a low UNK means the method mistakenly classifies a large number of samples from unknown classes as one of the known classes. This caveat cannot be captured by the combination of OS* and OS as the evaluation metrics.

On the more challenging Office-Home dataset, our method achieves the best HOS on 6 out of 12 tasks whilst the second-best method ROS \citep{bucci2020effectiveness} performs the best on 4 tasks. Again, our method achieves the best average HOS of 67.0\% followed by 66.2\% by \citet{bucci2020effectiveness} and 64.7\% by \citet{saito2018open}. The experimental results on both datasets are consistent in that methods achieving high OS* do not necessarily perform well in practice since they may make too many mistakes on the target samples from unknown classes as we can see from the results of UAN \cite{you2019universal} and PGL \cite{luo2020progressive}.

\subsection{Effect of Hyper-Parameters}
We investigate the effect of four hyper-parameters in our method: the dimensionality of the PCA subspace $d_{PCA}$, the dimensionality of the OSLPP subspace $d$, the number of initially rejected samples $n_r$ and the number of iterations $T$. To these ends, we set the investigated hyper-parameter to the values within a pre-defined set and the others fixed as the default in our experiments. The investigated values for four hyper-parameters are $d_{PCA} \in \{1024, 512, 256, 128, 64, 32, 16, 8\}$, $d \in \{512, 256, 128, 64, 32, 16, 8\}$, $n_r \in \{40, 60, 80, 100, 120, 140, 160, 180, 200, 220\}$ ($\times 10$ for the Office-Home dataset since this dataset contains much more samples) and $T \in \{6, 8, 10, 12, 14, 16, 18, 20\}$, respectively. 

\begin{figure*}
    \centering
    \includegraphics[width=0.95\textwidth]{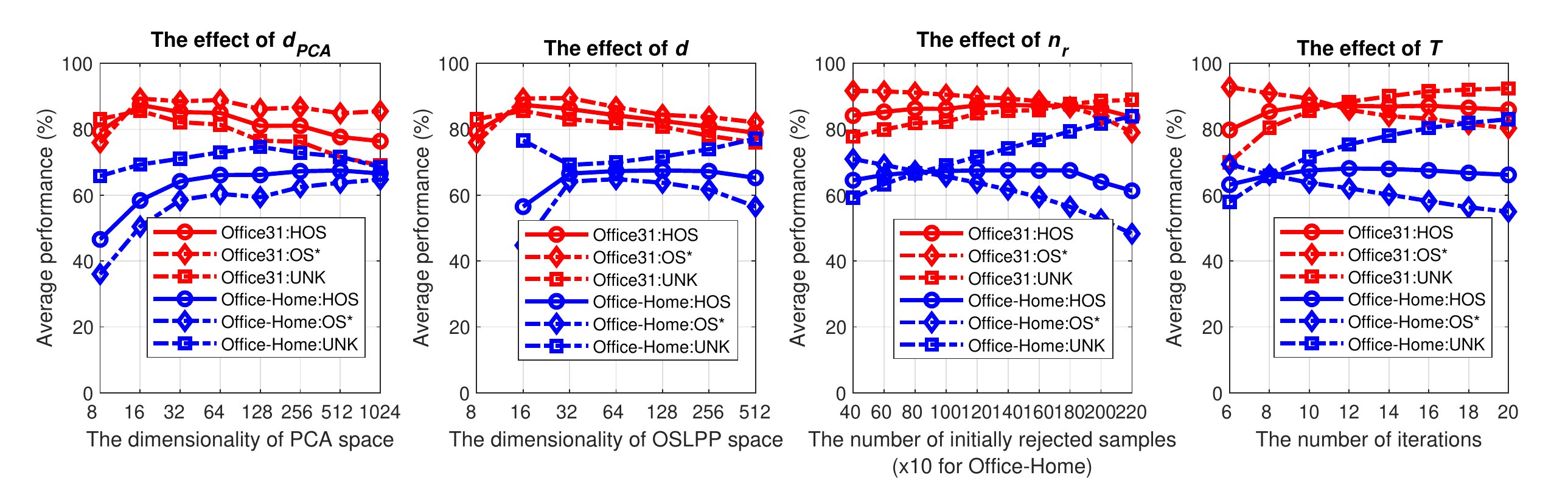}
    \caption{The effect of hyper-parameters.}
    \label{fig:hyper-parameters}
\end{figure*}

The results are shown in Figure \ref{fig:hyper-parameters}. We report the average OS*, UNK and HOS for each experiment. The optimal number of PCA dimensionality for Office31 is 16 and a smaller or greater value will lead to a slight performance drop for both OS* and UNK. This result means the ResNet50 features are discriminative enough to separate 10 shared known classes in this simple dataset so that the unsupervised PCA can extract the most useful information in the first 16 principal components whilst more dimensions hurt the performance slightly. Similarly, the optimal dimensionality of OSLPP subspace is 16 for Office31 and more dimensions lead to slightly worse performance. For the more challenging Office-Home dataset with 25 shared known classes and 40 unknown classes, the average HOS is less sensitive to the subspace dimensionality and optimal average HOS can be achieved within a large range of values for $d_{PCA}$ (i.e. 32-1024) and $d$ (i.e. 32-512). In addition, the values of $d_{PCA}$ and $d$ also affect the trade-off between the OS* and UNK although their harmonic mean HOS is marginally affected.

The effects of $n_r$ and $T$ are more understandable and consistent on the two datasets. On one hand, the performance of our method in terms of HOS is not sensitive to these two hyper-parameters given that the optimal HOS can be achieved with a large range of values for $n_r$ (i.e. 40-180 for Office31 and 400-1800 for Office-Home) and $T$ (i.e. 8-20).
On the other hand, both the $n_r$ and $T$ control the trade-off between the recognition accuracy of known classes OS* and the accuracy of unknown classes UNK. Increasing the number of initially rejected samples $n_r$ will lead to more samples recognised as from unknown classes hence an increased UNK. Increasing the number of iterations $T$ means selecting samples as known classes more slowly whilst the pace of rejecting samples as unknown classes is not affected. As a result, more samples will be rejected and recognised as unknown classes after more iterations hence a higher UNK can be achieved. Along with the improvement of UNK, our method suffers from the decrease of OS* although their harmonic mean HOS is stable.

Overall, our proposed method is not sensitive to hyper-parameters. In addition, we can trade off the recognition accuracy of known classes and unknown classes by adjusting the hyper-parameters of $n_r$ and $T$ according to the requirements in practice.

\subsection{Computational Complexity}\label{sec:complexity}
The complexity of PCA is $\mathcal{O}(d_0n^2+d_0^3)$. The complexity of OSLPP is $\mathcal{O}(2n^2d_{PCA}+d_{PCA}^3)$ which is repeated for $T$ times and leads to approximately $\mathcal{O}(T(2n^2d_{PCA}+d_{PCA}^3))$. In our experiments, we use a laptop with an Intel Core i5-7300HQ CPU and 32G memory RAM. Running 6 tasks of the Office31 dataset takes approximately 14 seconds and running 12 tasks of the Office-Home dataset takes approximately 10 minutes.

Regarding memory usage, when the number of samples $n=n_s+n_t$ is much greater than the dimensionality, the memory complexity is $\mathcal{O}(n^2)$. As a result, our method has the limitation of scaling up to the extremely large datasets (e.g., $n>100,000$) for which the neural networks based approaches can be better choices.

\subsection{Visualisation}
\begin{figure}
    \centering
    \includegraphics[width=0.45\textwidth]{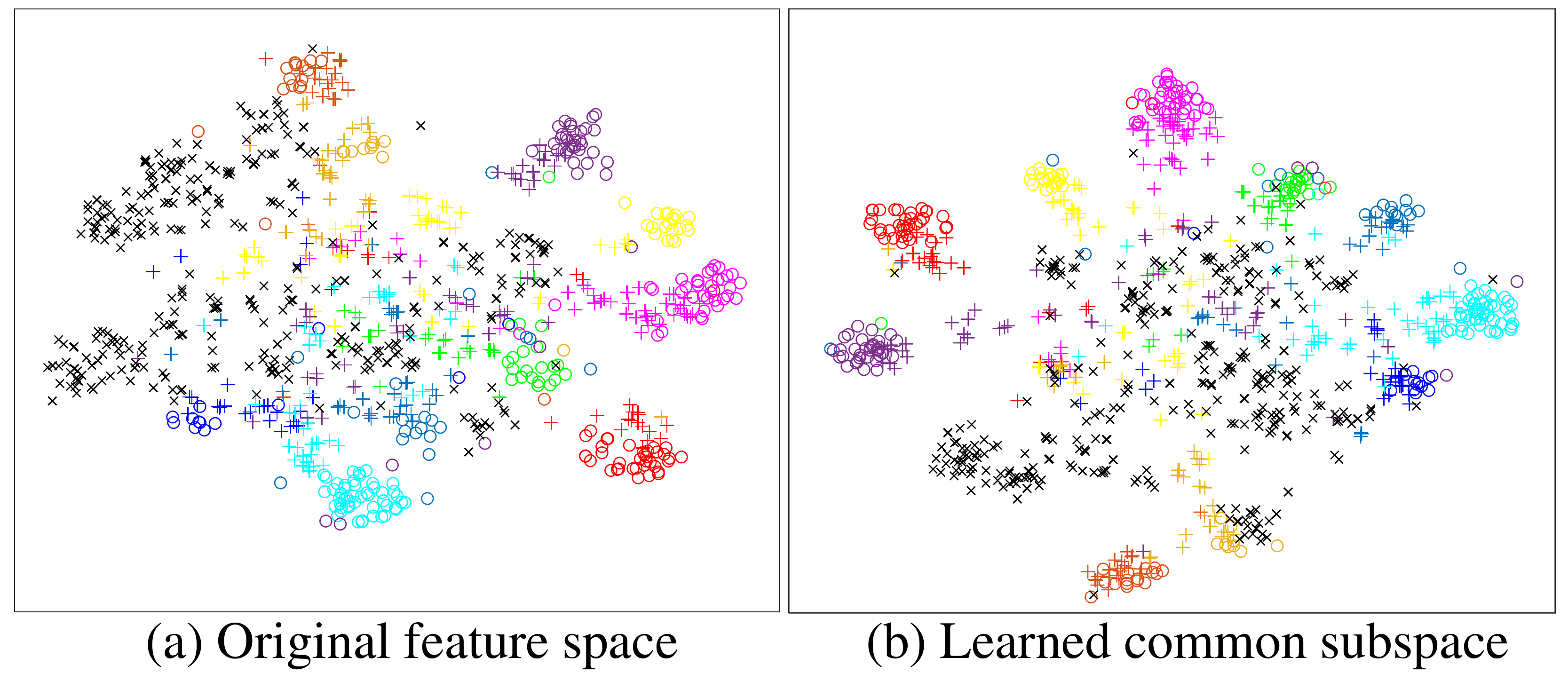}
    \caption{Feature visualisation via t-SNE. Source and target samples are represented by `o' and `+' symbols respectively; different known classes are denoted by colors and the unknown classes are denoted by black `$\times$'.}
    \label{fig:visualisation}
\end{figure}

We take the task Art$\to$Clipart as an example to inspect how the class separability can be improved in the learned subspace by visualising the features with the t-SNE technique. To avoid clutters, we select the first shared known classes and the first 10 unknown classes for visualisation. As shown in Figure \ref{fig:visualisation}, the samples from the source domain (`o') and the target domain (`+') in the learned common subspace are better aligned class-wisely whilst the unknown classes (`$\times$') are also more separated from the known classes.
\section{Discussion and Conclusion}
We address the OSDA problem in the image classification domain by proposing a novel OSLPP algorithm and a progressive pseudo-labelled samples selection and rejection strategy. The OSLPP adapts the original LPP algorithm to the OSDA scenario by considering the labelled source samples and pseudo-labelled target samples which have been either selected or rejected. Experimental results on two benchmark datasets demonstrate our proposed method can outperform state-of-the-art approaches to OSDA and is more efficient. The method is also insensitive to the hyper-parameters in terms of the harmonic mean but provides the flexibility of trading-off the accuracy of known and unknown classes which can be useful in real-world applications.

Our method suffers from the common issue of how to set proper hyper-parameters to adjust the recognition accuracy of known classes and unknown classes. In a real-world scenario, we do not have prior knowledge of how many unlabelled target samples are from unknown classes if there are any. In our method, this issue corresponds to the question of how to set the values of $n_r$ and $T$ given a domain adaptation task to achieve the best performance. This is also the key problem in Universal Domain Adaptation problems \cite{you2019universal} and will be left to the future work.

\bibliography{aaai22}

\begin{thebibliography}{24}
\providecommand{\natexlab}[1]{#1}

\bibitem[{Bucci, Loghmani, and Tommasi(2020)}]{bucci2020effectiveness}
Bucci, S.; Loghmani, M.~R.; and Tommasi, T. 2020.
\newblock On the effectiveness of image rotation for open set domain
  adaptation.
\newblock In \emph{European Conference on Computer Vision}, 422--438. Springer.

\bibitem[{Chen et~al.(2019{\natexlab{a}})Chen, Chen, Jiang, and
  Jin}]{chen2018joint}
Chen, C.; Chen, Z.; Jiang, B.; and Jin, X. 2019{\natexlab{a}}.
\newblock Joint Domain Alignment and Discriminative Feature Learning for
  Unsupervised Deep Domain Adaptation.
\newblock In \emph{AAAI Conference on Artificial Intelligence}.

\bibitem[{Chen et~al.(2019{\natexlab{b}})Chen, Xie, Huang, Rong, Ding, Huang,
  Xu, and Huang}]{chen2019progressive}
Chen, C.; Xie, W.; Huang, W.; Rong, Y.; Ding, X.; Huang, Y.; Xu, T.; and Huang,
  J. 2019{\natexlab{b}}.
\newblock Progressive Feature Alignment for Unsupervised Domain Adaptation.
\newblock In \emph{IEEE Conference on Computer Vision and Pattern Recognition},
  627--636.

\bibitem[{Deng et~al.(2009)Deng, Dong, Socher, Li, Li, and
  Fei-Fei}]{deng2009imagenet}
Deng, J.; Dong, W.; Socher, R.; Li, L.-J.; Li, K.; and Fei-Fei, L. 2009.
\newblock Image{N}et: A large-scale hierarchical image database.
\newblock In \emph{IEEE conference on computer vision and pattern recognition},
  248--255. {IEEE}.

\bibitem[{Fang et~al.(2020)Fang, Lu, Liu, Xuan, and Zhang}]{fang2020open}
Fang, Z.; Lu, J.; Liu, F.; Xuan, J.; and Zhang, G. 2020.
\newblock Open set domain adaptation: Theoretical bound and algorithm.
\newblock \emph{IEEE transactions on neural networks and learning systems}.

\bibitem[{Fu et~al.(2020)Fu, Cao, Long, and Wang}]{fu2020learning}
Fu, B.; Cao, Z.; Long, M.; and Wang, J. 2020.
\newblock Learning to detect open classes for universal domain adaptation.
\newblock In \emph{European Conference on Computer Vision}, 567--583. Springer.

\bibitem[{He et~al.(2016)He, Zhang, Ren, and Sun}]{he2016deep}
He, K.; Zhang, X.; Ren, S.; and Sun, J. 2016.
\newblock Deep residual learning for image recognition.
\newblock In \emph{IEEE conference on computer vision and pattern recognition},
  770--778.

\bibitem[{He and Niyogi(2004)}]{he2004locality}
He, X.; and Niyogi, P. 2004.
\newblock Locality preserving projections.
\newblock In \emph{Advances in neural information processing systems},
  153--160.

\bibitem[{Kundu et~al.(2020)Kundu, Venkat, Revanur, Babu
  et~al.}]{kundu2020towards}
Kundu, J.~N.; Venkat, N.; Revanur, A.; Babu, R.~V.; et~al. 2020.
\newblock Towards inheritable models for open-set domain adaptation.
\newblock In \emph{Proceedings of the IEEE/CVF Conference on Computer Vision
  and Pattern Recognition}, 12376--12385.

\bibitem[{Li et~al.(2021)Li, Kang, Zhu, Wei, and Yang}]{li2021domain}
Li, G.; Kang, G.; Zhu, Y.; Wei, Y.; and Yang, Y. 2021.
\newblock Domain Consensus Clustering for Universal Domain Adaptation.
\newblock In \emph{Proceedings of the IEEE/CVF Conference on Computer Vision
  and Pattern Recognition}, 9757--9766.

\bibitem[{Liu et~al.(2019)Liu, Cao, Long, Wang, and Yang}]{liu2019separate}
Liu, H.; Cao, Z.; Long, M.; Wang, J.; and Yang, Q. 2019.
\newblock Separate to adapt: Open set domain adaptation via progressive
  separation.
\newblock In \emph{Proceedings of the IEEE/CVF Conference on Computer Vision
  and Pattern Recognition}, 2927--2936.

\bibitem[{Luo et~al.(2020)Luo, Wang, Huang, and
  Baktashmotlagh}]{luo2020progressive}
Luo, Y.; Wang, Z.; Huang, Z.; and Baktashmotlagh, M. 2020.
\newblock Progressive graph learning for open-set domain adaptation.
\newblock In \emph{International Conference on Machine Learning}, 6468--6478.
  PMLR.

\bibitem[{Pan et~al.(2020)Pan, Yao, Li, Ngo, and Mei}]{pan2020exploring}
Pan, Y.; Yao, T.; Li, Y.; Ngo, C.-W.; and Mei, T. 2020.
\newblock Exploring category-agnostic clusters for open-set domain adaptation.
\newblock In \emph{Proceedings of the IEEE/CVF Conference on Computer Vision
  and Pattern Recognition}, 13867--13875.

\bibitem[{Panareda~Busto and Gall(2017)}]{panareda2017open}
Panareda~Busto, P.; and Gall, J. 2017.
\newblock Open set domain adaptation.
\newblock In \emph{Proceedings of the IEEE International Conference on Computer
  Vision}, 754--763.

\bibitem[{Saenko et~al.(2010)Saenko, Kulis, Fritz, and
  Darrell}]{saenko2010adapting}
Saenko, K.; Kulis, B.; Fritz, M.; and Darrell, T. 2010.
\newblock Adapting visual category models to new domains.
\newblock In \emph{European Conference on Computer Vision}, 213--226. Springer.

\bibitem[{Saito et~al.(2020)Saito, Kim, Sclaroff, and
  Saenko}]{saito2020universal}
Saito, K.; Kim, D.; Sclaroff, S.; and Saenko, K. 2020.
\newblock Universal Domain Adaptation through Self Supervision.
\newblock \emph{Advances in Neural Information Processing Systems}, 33.

\bibitem[{Saito et~al.(2018)Saito, Yamamoto, Ushiku, and
  Harada}]{saito2018open}
Saito, K.; Yamamoto, S.; Ushiku, Y.; and Harada, T. 2018.
\newblock Open set domain adaptation by backpropagation.
\newblock In \emph{Proceedings of the European Conference on Computer Vision
  (ECCV)}, 153--168.

\bibitem[{Venkateswara et~al.(2017)Venkateswara, Eusebio, Chakraborty, and
  Panchanathan}]{venkateswara2017deep}
Venkateswara, H.; Eusebio, J.; Chakraborty, S.; and Panchanathan, S. 2017.
\newblock Deep hashing network for unsupervised domain adaptation.
\newblock In \emph{IEEE Conference on Computer Vision and Pattern Recognition},
  5018--5027.

\bibitem[{Wang and Breckon(2020)}]{wang2020unsupervised}
Wang, Q.; and Breckon, T.~P. 2020.
\newblock Unsupervised domain adaptation via structured prediction based
  selective pseudo-labeling.
\newblock In \emph{AAAI Conference on Artificial Intelligence}.

\bibitem[{Wang and Breckon(2021)}]{wang2021source}
Wang, Q.; and Breckon, T.~P. 2021.
\newblock Source Class Selection With Label Propagation For Partial Domain
  Adaptation.
\newblock In \emph{2021 IEEE International Conference on Image Processing
  (ICIP)}, 769--773. IEEE.

\bibitem[{Wang and Breckon(2022)}]{wang2020generalized}
Wang, Q.; and Breckon, T.~P. 2022.
\newblock Generalized Zero-Shot Domain Adaptation via Coupled Conditional
  Variational Autoencoders.
\newblock \emph{Pattern Recognition}, 123.

\bibitem[{Wang, Bu, and Breckon(2019)}]{wang2019unifying}
Wang, Q.; Bu, P.; and Breckon, T.~P. 2019.
\newblock Unifying unsupervised domain adaptation and zero-shot visual
  recognition.
\newblock In \emph{International Joint Conference on Neural Networks}.

\bibitem[{Wold, Esbensen, and Geladi(1987)}]{wold1987principal}
Wold, S.; Esbensen, K.; and Geladi, P. 1987.
\newblock Principal component analysis.
\newblock \emph{Chemometrics and intelligent laboratory systems}, 2(1-3):
  37--52.

\bibitem[{You et~al.(2019)You, Long, Cao, Wang, and Jordan}]{you2019universal}
You, K.; Long, M.; Cao, Z.; Wang, J.; and Jordan, M.~I. 2019.
\newblock Universal domain adaptation.
\newblock In \emph{Proceedings of the IEEE/CVF conference on computer vision
  and pattern recognition}, 2720--2729.

\end{thebibliography}

\end{document}